\documentclass[11pt]{article}

\PassOptionsToPackage{table}{xcolor} 
\usepackage[preprint]{acl}

\usepackage{times}
\usepackage{latexsym}
\usepackage[T1]{fontenc}
\usepackage[utf8]{inputenc}
\usepackage{microtype}
\usepackage{inconsolata}
\usepackage{graphicx}

\usepackage{booktabs}
\usepackage{rotating} 
\usepackage{makecell}
\usepackage{amsmath}
\usepackage{algorithm}
\usepackage{algorithmic}
\usepackage{xcolor}
\usepackage{tikz}
\usetikzlibrary{positioning, arrows.meta, calc, fit, backgrounds}
\usepackage{pifont}

\definecolor{cpBlue}{HTML}{0072B2}
\definecolor{cpSky}{HTML}{56B4E9}
\definecolor{cpOrange}{HTML}{E69F00}
\definecolor{cpGreen}{HTML}{009E73}
\definecolor{cpPurple}{HTML}{CC79A7}
\definecolor{cpVermillion}{HTML}{D55E00}

\newcommand{\best}{\cellcolor{cpBlue!35}}
\newcommand{\second}{\cellcolor{cpBlue!12}}

\newif\ifreview \reviewtrue

\title{Beyond Ideological Similarity:~ Theme and Tone Supervision for Political Text Classification}

\author{Upasana Chatterjee \\
  Department of Computer Science \\
  Columbia University \\
  \texttt{uc2143@columbia.edu} \\\And
  Daniel Bauer \\
  Department of Computer Science \\
  Columbia University \\
  \texttt{bauer@cs.columbia.edu} \\}

\begin{document}

\maketitle

\begin{abstract}
Political-text classifiers support research on ideology, stance, and media bias, yet a high average score can hide errors that reverse a position or systematically erase and attribute bias. We test whether supervision from semantic themes and net evaluative tone changes a RoBERTa representation trained with ideology triplet loss and masked language modeling. 
Across facet-level ideology, sentence-level media bias, and target-conditioned stance, no auxiliary objective dominates. Instead, theme, tone, and joint supervision help different tasks and alter different error directions.
For example, joint supervision lowers lexical-bias erasure at \textit{The New York Times} from 52.7\% to 36.9\%, but raises Against-to-Favor stance reversals from 12.8\% to 15.2\%.
These effects change when auxiliary labels cover only part of the pre-training corpus. Metalinguistic supervision is therefore a design choice, not a universal upgrade: researchers should test which signal matches the construct and error costs of the intended analysis.
\end{abstract}

\section{Introduction}
\label{sec:intro}
Political-text classifiers turn articles and posts into evidence about ideology, stance, and media bias. Researchers use their outputs to compare outlets, issues, groups, and periods. The classifier therefore shapes the object it measures. In content moderation, uneven classification can suppress legitimate speech while leaving harmful material unaddressed; public reporting and a human-rights assessment of Meta document such failures and their unequal effects \citep{AP-News-2021,bsrmeta2022}. Hate-speech systems can also reproduce racial bias when annotators and models mistake dialect features for toxicity \citep{sap-etal-2019-risk}. These cases show that context-sensitive errors do not fall evenly across speakers or communities.

Additionally, content labels are not always universal. They may vary with language, culture, target, and annotator perspective \citep{plank-2022-problem}. Calling a Left-leaning article Right-leaning reverses the inferred position; calling it Center-leaning weakens the position without inverting it. A stance model can likewise reverse Favor and Against, while a bias model can erase annotated bias or attribute bias where none was marked. Such errors can alter claims about political actors and outlets even when aggregate accuracy changes little. A classifier should therefore be judged not only by how often it is wrong, but by which direction, group, and construct bear the error.

\begin{figure*}[t]
\centering
\begin{tabular}{@{}p{0.48\textwidth}@{\hspace{0.04\textwidth}}p{0.48\textwidth}@{}}
\multicolumn{2}{@{}l@{}}{\textbf{Target-conditioned stance}}\\[\smallskipamount]
\fcolorbox{black!25}{black!5}{\parbox{0.94\linewidth}{
``It's a problem when explaining feminism, even in a calm \& complex level, cannot be understood. \#feminist \#respect \#EqualityForAll''\par\medskip
\textbf{RoBERTa:}~ \textcolor{cpOrange!85!black}{Against}\hfill
\textbf{All CPT arms:}~ \textcolor{cpBlue!80!black}{Favor}}}
&
\fcolorbox{black!25}{black!5}{\parbox{0.94\linewidth}{
``Helping getting laws for both patents to get paid maternity leave???!! WHEN WILL THIS MADNESS STOP?!?!?? \#feminismiscruelty''\par\medskip
\textbf{RoBERTa:}~ \textcolor{cpOrange!85!black}{Favor}\hfill
\textbf{All CPT arms:}~ \textcolor{cpBlue!80!black}{Against}}}
\end{tabular}
\caption{Examples of reversals corrected by continued pre-training. Both panels contain SemEval stance examples \citep{mohammad-etal-2016-semeval}. The examples motivate direct analysis of error direction rather than accuracy alone. \textcolor{cpOrange!85!black}{Orange} marks incorrect and \textcolor{cpBlue!80!black}{blue} correct predictions.}
\label{fig:motivating-reversals}
\end{figure*}

Figure~\ref{fig:motivating-reversals} illustrates the problem. The political direction of the tweets is apparent to a reader from the target, irony, and surrounding proposition, yet vanilla RoBERTa predicts the opposite stance. All four continued-pre-training arms predict the gold stance in both examples. This motivates a broader question: which signals help a model preserve political direction when the form of expression changes?

We study two metalinguistic signals: theme and tone. The Global Database of Events, Language, and Tone (GDELT) supplies semantic themes and a net-tone score for news \citep{leetaru2013gdelt}. Themes approximate what an article discusses; net tone approximates how it evaluates that content. Neither directly labels ideology, stance, or bias. We add these objectives to a backbone trained with triplet-loss pre-training (TLP), which organizes ideological similarity, and masked language modeling (MLM), which adapts the encoder to the corpus. Prior work links topic and sentiment to political language \citep{mullen2006preliminary,smirnova2017ideology,bhatia2018topic}, but their value as auxiliary signals depends on the task.

We ask three questions. RQ1: Does theme or tone supervision improve primary scores over an MLM+TLP backbone? RQ2: How does it change directional, outlet, and facet errors? RQ3: Does the pattern change when auxiliary labels cover only part of the pre-training corpus?

We make three contributions:
\begin{enumerate}
    \item We compare theme and tone supervision across four political-text constructs on a fixed contrastive backbone.
    \item We provide the first controlled political-text comparison, to our knowledge, of the same auxiliary targets under dense and naturally partial coverage.
    \item We evaluate direct, task-specific high-cost errors: Left--Right and Favor--Against reversals, bias erasure and attribution, and facet-level ideological errors.
\end{enumerate}

The results show that no single auxiliary objective is universally effective.
Theme, tone, and joint supervision reduce bias erasure or attribution for particular outlets and improve different ideological facets or stance targets.
Yet no head helps every task, and partial supervision changes several rankings. Political classifiers should therefore treat metalinguistic supervision as a design choice and audit it against the construct and error costs of the intended analysis.

\section{Related Work}
\label{sec:related_work}

\paragraph{Political classifiers are measurement instruments.}
Political classifiers assign ideology, stance, or bias to different units, from phrases and articles to users and outlets \citep{iyyer-etal-2014-political,preotiuc-pietro-etal-2017-beyond,baly-etal-2018-predicting}. Those units are not interchangeable. Sentiment and stance are distinct constructs \citep{bestvater2023sentiment}; bias at the article, paragraph, and sentence levels does not reduce to one quantity \citep{chen2020analyzing}; and outlet estimates aggregate lower-level predictions \citep{ronnback2025biasoutlets}. Recent work likewise shows that ideology proxies vary in bias and transfer \citep{ram2025practical}, stance performance changes with dataset and target ambiguity \citep{ng2025stancebias}, and domain-specific data can matter as much as architecture \citep{schneider2026german}. We therefore treat the training signal, its coverage, and the costly error type as parts of measurement design.

\paragraph{Relational political representation learning.}
Triplet and contrastive objectives organize political similarity while controlling topic or outlet cues. \citet{baly2020we} construct ideology triplets and evaluate article classification on unseen media. POLITICS compares ideologically distinct reports of the same story and adds a story objective and specialized masked language modeling \citep{liu2022politics}. MITweet and concept-flow models instead represent ideology across named facets \citep{liu2023MITweet,liu-etal-2024-encoding}, while BASIL separates lexical from informational bias \citep{fan2019basil}. We ask what weak presentation-level supervision adds after a strong MLM+TLP backbone has already learned relational political structure.

\paragraph{Why theme and tone.}
Political communication distinguishes what a text makes salient from how it evaluates that content. Framing and agenda-setting research links political meaning to issue selection, emphasis, and characterization \citep{entman1993framing,field-etal-2018-framing}, while media-bias research separates selection from evaluative statements \citep{dalessio2000media}. This literature motivates GDELT themes as weak supervision over topical and semantic content, not as frame labels \citep{leetaru2013gdelt,otmakhova-etal-2024-media}.

Tone captures a different signal. Stance names a position toward a target, whereas sentiment names evaluative polarity; the two diverge when a text evaluates another entity or when the same sentiment supports different positions \citep{sobhani-etal-2016-detecting,mohammad-etal-2016-semeval}. Biased wording also uses presupposition, hedging, factivity, and intensification beyond positive--negative sentiment \citep{recasens2013weasel}. These findings motivate net tone for target-centered stance and lexical bias, while warning that it is not a stable proxy for left--right ideology.

\paragraph{Conditional transfer and label coverage.}
Auxiliary tasks help only when they induce useful representations; related tasks can produce gains or interference \citep{bingel-sogaard-2017-identifying,pruksachatkun-etal-2020-intermediate}. Prior political-text studies vary objectives, weak labels, or training scale, but do not compare the same auxiliary targets under both dense and naturally partial coverage while holding a triplet-based contrastive backbone fixed. Our design tests where theme and tone help, how they reshape task-specific errors, and whether the pattern changes when only the metadata-matched portion of the corpus supplies auxiliary gradients.

\section{Experimental Design}
\label{sec:experimental_design}

\subsection{Downstream tasks}
Table~\ref{tab:task-design} summarizes the three downstream tasks, and stage~3 of Figure~\ref{fig:experimental-design} places them in the pipeline. AllSides motivates the analysis and selects the pre-training configuration. MITweet, BASIL, and SemEval use that locked configuration on different constructs and units\footnote{Our code is released under the MIT License; available at \url{https://github.com/upasanachatterjee/metalinguistic_features_in_political_bias_detection}.}.

\begin{table}[t]
\centering
\setlength{\tabcolsep}{3pt}
\resizebox{\linewidth}{!}{%
\begin{tabular}{@{}llll@{}}
\toprule
Dataset & Unit / Label & Data Splits & Severe Errors \\
\midrule
MITweet & \makecell[l]{tweet--facet \\ L/C/R} & \makecell[l]{11,992/2,571/2,542 \\ 4 held-out folds} & \makecell[l]{reversal; \\ F1 by facet} \\
\addlinespace
BASIL & \makecell[l]{sentence \\ biased/ \\ not biased} & \makecell[l]{6,826/758/400 \\ 10 folds} & \makecell[l]{erasure/ \\ attribution \\ by outlet} \\
\addlinespace
SemEval & \makecell[l]{tweet--target \\ against/ \\ none/favor} & \makecell[l]{A: 2,814/100/1,249 \\ B: 2,814/100/707} & \makecell[l]{Against$\leftrightarrow$ \\ Favor} \\
\bottomrule
\end{tabular}%
}
\caption{Downstream tasks. Data Splits give training, validation, and test counts, with fold counts noted where used. L/C/R denotes Left, Center, and Right.}
\label{tab:task-design}
\end{table}

\paragraph{AllSides.}
We curate the article-level AllSides corpus released by \citet{baly2020we} to balance Left, Center, and Right labels and equalize topic coverage. The labels are article annotations, not outlet-level labels, and the downstream split is source-disjoint: test publishers appear neither in fine-tuning training nor continued pre-training. Because the test labels informed the triplet count, epoch, and initial analysis, all AllSides findings remain exploratory. This dataset is available under the Apache-2.0 license\footnote{The source-disjoint split used here is available at \url{https://huggingface.co/datasets/upasanachatterjee/AllSides-media-split}; we also release a random split at \url{https://huggingface.co/datasets/upasanachatterjee/AllSides-random-split}.}.

\paragraph{MITweet.}
MITweet assigns ideological direction to tweets within twelve political facets \citep{liu2023MITweet}. We evaluate a random ideology split and four round-robin held-out-facet folds, each withholding a disjoint subset so that every available facet is evaluated as held out. We report task- and facet-level effects because one mean can hide gains on one political dimension and losses on another. Its facets can draw on both topical and evaluative signals. We do not evaluate the separate relevance task. This dataset is available under the GNU GPL V3.0 license.

\paragraph{BASIL.}
BASIL marks lexical and informational bias in matched-event news \citep{fan2019basil}. We cast each annotation type as binary sentence classification and use ten folds, with positive-class F1 as the primary score. We also report bias erasure and attribution for Fox News, HuffPost, and \textit{The New York Times}, because the same average can mask different outlet effects. This dataset is available under the CC-BY-4.0 license.

\paragraph{SemEval-2016 Task 6.}
SemEval labels stance toward a named target as Against, None, or Favor \citep{mohammad-etal-2016-semeval}. We hold out 100 training examples for validation. Task A evaluates targets seen during training; Task B follows the official protocol and holds out Donald Trump. We include the target in the input because stance depends on the object of evaluation. This dataset is free for research use\footnote{See \url{https://www.saifmohammad.com/WebPages/StanceDataset.htm}.}.

\subsection{Continued pre-training}
\label{sec:continued_pretraining}

\begin{figure*}[t]
\centering
\resizebox{\textwidth}{!}{%
\begin{tikzpicture}[
    box/.style={draw, thick, rounded corners, align=center, fill=white,
                inner sep=3pt},
    corpus/.style={box, text width=2.6cm, draw=cpBlue, fill=cpSky!20},
    enc/.style={box, text width=2.3cm, minimum height=2.5cm, draw=cpBlue, fill=cpBlue!25},
    head/.style={box, text width=5.3cm, inner ysep=2pt, draw=cpBlue, fill=cpSky!15},
    headdash/.style={head, dashed, draw=cpPurple, fill=cpPurple!15},
    data/.style={box, text width=11.0cm, draw=cpOrange!80!black, fill=cpOrange!15},
    panel/.style={box, minimum width=4.0cm, minimum height=1.7cm, draw=cpOrange!80!black},
    lock/.style={box, text width=2.2cm, minimum height=1.7cm, draw=cpGreen, fill=cpGreen!25},
    task/.style={box, text width=4.3cm, draw=cpGreen, fill=cpGreen!12},
    cell/.style={draw=gray!70, minimum width=0.9cm, minimum height=0.5cm,
                 inner sep=0pt},
    chosen/.style={cell, draw=cpOrange!80!black, thick, fill=cpOrange!55},
    stage/.style={font=\bfseries, anchor=west},
    note/.style={font=\itshape, align=left, text width=4.2cm},
    arr/.style={-Latex, thick}
  ]

  \begin{scope}[on background layer]
    \fill[cpBlue!6, rounded corners=3pt] (0.00,3.95) rectangle (17.60,7.30);
    \fill[cpOrange!7, rounded corners=3pt] (0.00,-0.10) rectangle (12.20,3.70);
    \fill[cpGreen!7, rounded corners=3pt] (12.50,-0.10) rectangle (17.60,3.70);
  \end{scope}
  \node[stage] at (0.10,7.05) {1.\ \ Continued pre-training};
  \node[stage] at (0.10,3.42) {2.\ \ Sequential configuration search};
  \node[stage] at (12.60,3.42) {3.\ \ Transfer};

  \node[corpus]        (dense) at (1.60,6.00) {\textbf{Dense subset}\\339K matched};
  \node[corpus,dashed] (fullc) at (1.60,4.85) {\textbf{Full corpus}\\1.03M articles};

  \node[enc] (rob) at (4.90,5.425) {RoBERTa-base\\encoder};

  \draw[arr] (dense.east) -- (rob.west|-dense);
  \draw[arr,dashed] (fullc.east) -- (rob.west|-fullc);

  \node[head]     (trip)  at (9.90,6.43) {\textbf{Triplet ($T_{id}$)} $\cdot$ \textit{POLITICS corpus}};
  \node[head]     (mlm)   at (9.90,5.77) {\textbf{MLM} $\cdot$ \textit{POLITICS corpus}};
  \node[headdash] (theme) at (9.90,5.11) {\textbf{Theme ($T_h$)} $\cdot$ \textit{GDELT matched}};
  \node[headdash] (tone)  at (9.90,4.45) {\textbf{Tone ($T_o$)} $\cdot$ \textit{GDELT matched}};

  \foreach \h in {trip,mlm,theme,tone}
    \draw[arr] (rob.east|-\h) -- (\h.west);

  \node[note, anchor=west] at (13.00,5.425)
    {Full corpus: TLP and MLM on all articles; Theme and Tone on the matched third only.};

  \draw[arr] (9.90,4.12) -- (9.90,3.05);
  \node[anchor=west] at (10.00,3.45) {4 arms};

  \node[data] (allsides) at (6.10,2.60)
    {\textbf{AllSides} \textit{(selection benchmark)}: article ideology (L/C/R);
     fine-tune, predict, and take the best macro-F1};

  \node[panel] (stepa) at (2.15,0.95) {};
  \node[anchor=north, align=center] at (2.15,1.85) {\textbf{Step 1: triplet count}\\at 1 epoch};
  \node[cell]   at (1.25,0.45) {1};
  \node[chosen] at (2.15,0.45) {16};
  \node[cell]   at (3.05,0.45) {32};

  \node[panel] (stepb) at (6.85,0.95) {};
  \node[anchor=north, align=center] at (6.85,1.85) {\textbf{Step 2: epochs}\\at 16 triplets};
  \node[chosen] at (6.40,0.45) {1};
  \node[cell]   at (7.30,0.45) {2};

  \node[lock] (locked) at (10.60,0.95)
    {\textbf{Final}\\16 triplets\\1 epoch};

  \draw[arr] (stepa.east) -- (stepb.west);
  \draw[arr] (stepb.east) -- (locked.west);

  \draw[arr] (2.15,0 |- allsides.south) -- (stepa.north);
  \draw[arr] (6.85,0 |- allsides.south) -- (stepb.north);

  \node[task] (mit) at (15.10,2.45) {\textbf{MITweet}\\facet ideology (L/C/R)};
  \node[task] (bas) at (15.10,1.45) {\textbf{BASIL}\\sentence bias (binary)};
  \node[task] (sem) at (15.10,0.45) {\textbf{SemEval-2016 Task 6}\\target stance (3-way)};

  \draw[thick] (locked.east) -- (12.35,0.95);
  \draw[thick] (12.35,0.45) -- (12.35,2.45);
  \foreach \t in {mit,bas,sem}
    \draw[arr] (12.35,0 |- \t.west) -- (\t.west);

\end{tikzpicture}%
}
\caption{Full experimental design. Four arms, obtained by toggling the two ablated heads, share a RoBERTa-base encoder and are pre-trained on the dense subset and on the full corpus. Two sequential AllSides searches then fix the triplet count and the epoch, and the locked configuration transfers unchanged to MITweet, BASIL, and SemEval-2016 Task~6. Dashed outlines mark the ablated theme and tone heads and the partially labeled full corpus; shaded cells mark the selected setting.}
\label{fig:experimental-design}
\end{figure*}
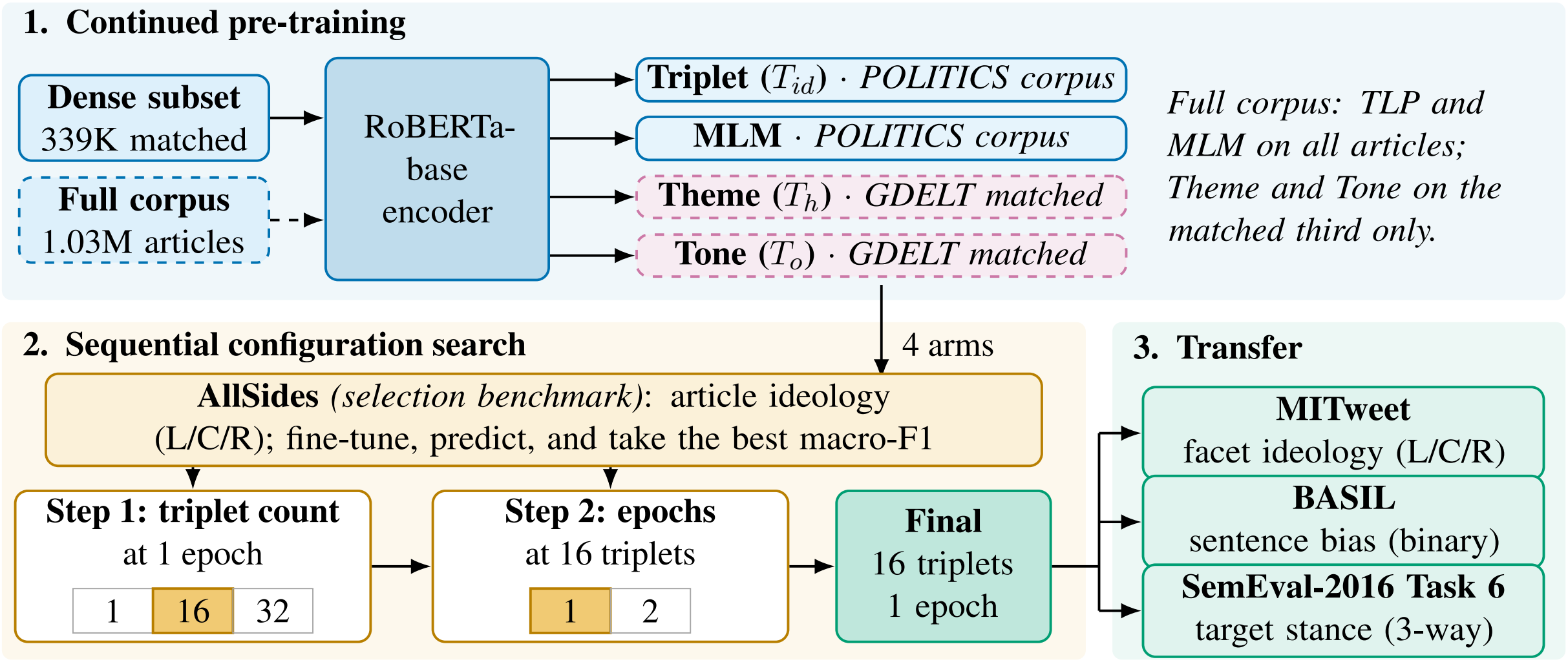

\paragraph{Corpus and supervision regimes.}
We use \textsc{BigNewsAlign}, the news corpus assembled for POLITICS \citep{liu2022politics}. The archive contains 1,025,180 unique articles available under the CC BY-NC-SA 4.0 license. We match 339,215 articles to non-empty GDELT theme and tone fields\footnote{The corpus with matched GDELT fields is available at \url{https://huggingface.co/datasets/upasanachatterjee/bignewsalign-with-gdelt}.}. GDELT releases its data under a permissive license\footnote{See details in their terms of use: \url{https://gdeltproject.org/about.html\#termsofuse}.}. This matched third forms the \emph{dense subset}. Every article in this subset can receive triplet-loss pre-training, masked-language modeling, theme, and tone supervision. Because every factorial arm sees the same texts, the dense subset provides the main controlled comparison.

In the full corpus, every eligible article contributes to TLP and MLM, while only matched articles contribute theme or tone loss. Explicit validity masks give missing labels zero gradient and distinguish them from valid all-zero theme vectors. Because the matched third is not a random sample (GDELT coverage reflects its collection window and the article-matching process) and the regimes also differ in size, dates, sources, and topics, their contrast shows how observed metadata coverage changes the training pattern, not the isolated effect of coverage.

GDELT \textsc{V2Themes} provides article-level theme codes; we retain the most frequent 2,000 themes and cast prediction as multi-label classification. From the several fields of \textsc{V2Tone} we use aggregate net tone as a scalar regression target.

\paragraph{Objectives.}
Let $f(x)$ denote the final-layer \texttt{[CLS]} representation of article $x$. Within each shuffled batch, an anchor $a$ and positive $p$ are distinct articles with the same Left or Right label; a negative $n$ has the opposite label. Center articles do not enter the triplet objective. For sampled triplets $\mathcal{T}$,
\begin{equation}
\begin{split}
\mathcal{L}_{\mathrm{TLP}}
={}& \frac{1}{|\mathcal{T}|}
\sum_{(a,p,n)\in\mathcal{T}}
\max\{d(f(a),f(p)) \\
&\quad -d(f(a),f(n))+m,0\},
\end{split}
\end{equation}
where $d$ is Euclidean distance and $m=1$. The loss is mean-reduced across sampled triplets.

The theme head predicts a 2,000-dimensional binary vector with binary cross-entropy, the tone head predicts net tone with mean-squared error, and MLM uses token-level cross-entropy. So for article $i$ the objective is:
\begin{equation}
\begin{split}
\mathcal{L}_{i}={}&
\lambda_{t}\mathcal{L}_{\mathrm{TLP},i}
+\lambda_{m}\mathcal{L}_{\mathrm{MLM},i} \\
&+\mathbf{1}(h,i)\lambda_{h}\mathcal{L}_{h,i}
+\mathbf{1}(o,i)\lambda_{o}\mathcal{L}_{o,i},
\end{split}
\end{equation}
where $\mathbf{1}(h,i)$ and $\mathbf{1}(o,i)$ indicate the presence of valid theme and tone labels for article $i$. All final runs use $\lambda_t=0.30$, $\lambda_h=2.45$, $\lambda_o=0.20$, and $\lambda_m=1.0$. We keep these static weights across arms and triplet counts. Appendix~\ref{appendix:weight-selection} records the gradient-based selection procedure.

\paragraph{Factorial design and configuration choice.}
Figure~\ref{fig:experimental-design} summarizes the design. During configuration search, $X\in\{1,16,32\}$ denotes the number of triplets per batch. The final matched arms use $X=16$:
\begin{itemize}
  \item MLM+TLP only ($MLM+T_{id}(16)$),
  \item MLM+TLP+Theme ($MLM+T_{id}(16)+T_h$),
  \item MLM+TLP+Tone ($MLM+T_{id}(16)+T_o$), and
  \item all objectives ($MLM+T_{id}(16)+T_h+T_o$).
\end{itemize}

AllSides selected 16 triplets per batch and 1 pre-training epoch (Appendix~\ref{appendix:pretraining}). We froze both choices, and the static loss weights, for MITweet, BASIL, SemEval, and the full-corpus runs (Appendix~\ref{app:training_configurations}).

\subsection{Baselines and fine-tuning}
The principal baseline is MLM+TLP trained on the same corpus as the auxiliary arms. RoBERTa measures the effect of continued pre-training.

We fine-tune each checkpoint under the same task-specific procedure. 
AllSides, MITweet, and SemEval use 5 seeds, except dense-subset SemEval Task A, which uses 3; BASIL uses 1 seed in each of 10 folds. Appendix~\ref{app:training_configurations} lists the optimizer, learning rate, batch sizes, early stopping, hardware, and seed set.

\subsection{Metrics and paired comparisons}
We report macro-F1 for AllSides and MITweet, positive-class F1 for BASIL, and the official mean of Against and Favor F1 for SemEval stance. In every performance table, bold marks the best value in each row and underlining marks the second best. The controlled contrasts compare Theme, Tone, and Theme+Tone with MLM+TLP.

\begin{table*}[th!]
\centering
\setlength{\tabcolsep}{4.5pt}
\resizebox{\linewidth}{!}{%
\begin{tabular}{llccccc}
\toprule
 &  &  &  & \multicolumn{3}{c}{Metalinguistic supervision} \\
\cmidrule(lr){5-7}
Dataset & Task and criterion & RoBERTa & $MLM+T_{id}(16)$ & $+T_h$ & $+T_o$ & $+T_h+T_o$ \\
\midrule
MITweet & Ideology F1, random $\uparrow$ & $50.7_{0.7}$ & $50.1_{0.5}$ & $50.1_{0.7}$ & \best$\mathbf{51.1_{0.5}}$ & \second$\underline{50.9_{0.7}}$ \\
 & Ideology reversal, random $\downarrow$ & $33.1_{2.1}$ & \second$\underline{32.7_{1.0}}$ & \best$\mathbf{32.4_{1.3}}$ & \best$\mathbf{32.4_{0.8}}$ & $32.8_{1.5}$ \\
 & Ideology F1, held-out facet $\uparrow$ & $27.5_{7.9}$ & \best$\mathbf{28.1_{7.7}}$ & $27.6_{7.5}$ & \second$\underline{27.9_{7.9}}$ & \second$\underline{27.9_{7.8}}$ \\
 & Ideology reversal, held-out facet $\downarrow$ & $62.3_{16.0}$ & \best$\mathbf{60.6_{16.4}}$ & \second$\underline{61.8_{16.1}}$ & \second$\underline{61.8_{17.1}}$ & \second$\underline{61.8_{16.6}}$ \\
\midrule
BASIL & Informational bias F1 $\uparrow$ & $45.0_{4.5}$ & \best$\mathbf{45.6_{3.3}}$ & $44.1_{3.5}$ & \second$\underline{45.5_{4.2}}$ & $42.1_{5.4}$ \\
 & Informational erasure $\downarrow$ & \second$\underline{38.9_{6.2}}$ & \best$\mathbf{38.0_{7.4}}$ & $41.0_{8.2}$ & $41.0_{6.6}$ & $43.3_{7.9}$ \\
 & Informational attribution $\downarrow$ & $20.2_{5.6}$ & $19.7_{3.3}$ & \second$\underline{19.5_{4.3}}$ & \best$\mathbf{18.2_{3.7}}$ & $20.4_{3.7}$ \\
 & Lexical bias F1 $\uparrow$ & $28.4_{5.6}$ & $26.8_{8.5}$ & $28.5_{7.1}$ & \best$\mathbf{30.0_{5.6}}$ & \second$\underline{29.1_{4.0}}$ \\
 & Lexical erasure $\downarrow$ & $46.8_{14.7}$ & $51.4_{18.9}$ & $50.0_{15.5}$ & \second$\underline{43.6_{9.6}}$ & \best$\mathbf{39.1_{8.6}}$ \\
 & Lexical attribution $\downarrow$ & $12.8_{3.3}$ & \second$\underline{12.2_{4.6}}$ & \best$\mathbf{12.1_{6.2}}$ & $13.4_{4.7}$ & $15.4_{4.7}$ \\
\midrule
SemEval & Stance F1, seen target $\uparrow$ & $70.8_{0.7}$ & $70.3_{1.5}$ & $58.4_{19.1}$ & \second$\underline{71.1_{1.1}}$ & \best$\mathbf{71.5_{1.7}}$ \\
 & Stance reversal, seen target $\downarrow$ & $18.6_{0.8}$ & \best$\mathbf{17.7_{1.1}}$ & $22.1_{6.8}$ & $18.7_{0.7}$ & \second$\underline{17.8_{0.8}}$ \\
 & Stance F1, held-out target $\uparrow$ & $19.9_{6.9}$ & \best$\mathbf{34.5_{7.4}}$ & $31.0_{8.8}$ & \second$\underline{31.2_{1.4}}$ & $30.8_{8.2}$ \\
 & Stance reversal, held-out target $\downarrow$ & \best$\mathbf{9.8_{3.7}}$ & $18.8_{5.0}$ & $24.6_{9.9}$ & \second$\underline{14.5_{2.3}}$ & $15.3_{3.2}$ \\
\bottomrule
\end{tabular}%
}
\caption{Dense-subset results at the selected configuration, in percent. Each task contributes its primary score and the directional error rate that would distort the corresponding measurement. MITweet reports macro-F1, BASIL reports positive-class F1, and SemEval reports the mean of Against and Favor F1; reversal is endpoint-to-endpoint error, erasure is the false-negative rate, and attribution is the false-positive rate. Arrows give the preferred direction. Bold (\textcolor{cpBlue}{dark blue}) marks the best and underline (\textcolor{cpBlue!60}{light blue}) the second-best distinct value in each row. Appendix~\ref{appendix:statistics} gives the paired intervals. Subscripts give standard deviations across five fine-tuning seeds; for SemEval Task A, across three; for BASIL, across ten folds with one seed each; and for held-out-facet MITweet, across all 20 fold--seed runs.}
\label{tab:main-dense}
\end{table*}

For ordered three-class tasks, a reversal maps one endpoint to the other. For endpoint labels $E=\{0,2\}$, the reversal rate is the share of endpoint examples predicted as the opposite endpoint, $\sum_i \mathbf{1}[y_i\in E,\ \hat{y}_i=2-y_i] / \sum_i \mathbf{1}[y_i\in E]$.
For BASIL, \emph{erasure} is the false-negative rate and \emph{attribution} the false-positive rate, which we report by outlet. We also report the Fox--\textit{Times} erasure gap, because unequal error between outlets can bias a comparative claim even when average error is acceptable.

For AllSides, MITweet, and SemEval, we report the paired mean difference from MLM+TLP and a 95\% paired $t$ interval over matched fine-tuning seeds. Held-out-facet MITweet is averaged over folds within each seed. BASIL intervals summarize matched folds descriptively because their training sets overlap. All intervals condition on one continued-pre-training checkpoint per arm and omit pre-training variance. Appendix~\ref{appendix:statistics} gives the paired summaries.

\section{Results}
\label{sec:results}

AllSides served as the selection benchmark for the pre-training checkpoint used in fine-tuning (Section~\ref{sec:experimental_design}), so we report its results in Appendix~\ref{appendix:results} and exclude it from the main analysis.

\subsection{Auxiliary supervision transfers selectively}
Table~\ref{tab:main-dense} reports the dense-subset factorial at the selected configuration. No auxiliary objective leads every task.
(RQ1)

A metalinguistic arm has the highest mean on three of six primary scores: Tone on random-split MITweet and BASIL lexical bias, and joint supervision on seen-target SemEval. MLM+TLP has the highest mean on held-out MITweet facets, BASIL informational bias, and held-out-target SemEval. However, no paired 95\% interval for an auxiliary gain excludes zero, and the only interval that does is a 3.5-point loss for joint supervision on BASIL informational bias (Appendix~\ref{appendix:statistics}).

Theme and tone do not supply a common increment to political classification. Their value changes with the construct, unit, and train--test relation. Table~\ref{tab:selected-gains} gives representative subgroup results. Theme improves MITweet Political Regime, Tone improves Diplomatic Strategy, and joint supervision improves Military Force. On seen-target SemEval, joint supervision improves all five targets; its largest gains are on Abortion, Atheism, and Hillary Clinton. These gains weaken under held-out facets and targets, where the relational backbone remains strongest overall.

\begin{table}[t]
\centering
\small
\setlength{\tabcolsep}{2.6pt}
\begin{tabular}{@{}lrrr@{}}
\toprule
\makecell[tl]{Example\\(selected head)} & Backbone & Aux. & $\Delta$ \\
\midrule
\multicolumn{4}{@{}l}{\textit{F1 (higher is better)}} \\
\makecell[tl]{MITweet: Political Regime\\($+T_h$)} & $\underline{35.7}$ & $\mathbf{36.7}$ & \cellcolor{cpBlue!30}$+1.0$ \\
\makecell[tl]{MITweet: Dipl. Strategy\\($+T_o$)} & $\underline{42.8}$ & $\mathbf{45.2}$ & \cellcolor{cpBlue!30}$+2.4$ \\
\makecell[tl]{MITweet: Military Force\\($+T_h+T_o$)} & $\underline{47.5}$ & $\mathbf{51.9}$ & \cellcolor{cpBlue!50}$+4.4$ \\
\midrule
\makecell[tl]{SemEval: Abortion\\($+T_h+T_o$)} & $\underline{53.8}$ & $\mathbf{59.0}$ & \cellcolor{cpBlue!50}$+5.2$ \\
\makecell[tl]{SemEval: Atheism\\($+T_h+T_o$)} & $\underline{56.9}$ & $\mathbf{61.0}$ & \cellcolor{cpBlue!50}$+4.1$ \\
\makecell[tl]{SemEval: Hillary Clinton\\($+T_h+T_o$)} & $\underline{68.3}$ & $\mathbf{71.4}$ & \cellcolor{cpBlue!50}$+3.1$ \\
\makecell[tl]{SemEval: Climate\\($+T_h+T_o$)} & $\underline{44.7}$ & $\mathbf{45.2}$ & \cellcolor{cpBlue!15}$+0.5$ \\
\makecell[tl]{SemEval: Feminism\\($+T_h+T_o$)} & $\underline{58.0}$ & $\mathbf{58.4}$ & \cellcolor{cpBlue!15}$+0.4$ \\
\midrule
\multicolumn{4}{@{}l}{\textit{Error rate (lower is better)}} \\
\makecell[tl]{BASIL Fox erasure\\($+T_h+T_o$)} & $\underline{51.6}$ & $\mathbf{45.4}$ & \cellcolor{cpBlue!50}$-6.2$ \\
\makecell[tl]{BASIL NYT erasure\\($+T_h+T_o$)} & $\underline{52.7}$ & $\mathbf{36.9}$ & \cellcolor{cpBlue!50}$-15.8$ \\
\midrule
\makecell[tl]{SemEval F$\rightarrow$A\\($+T_h+T_o$)} & $\underline{29.3}$ & $\mathbf{24.1}$ & \cellcolor{cpBlue!50}$-5.2$ \\
\makecell[tl]{SemEval A$\rightarrow$F\\($+T_h+T_o$)} & $\mathbf{12.8}$ & $\underline{15.2}$ & \cellcolor{cpOrange!30}$+2.4$ \\
\makecell[tl]{SemEval Trump reversal\\($+T_o$)} & $\underline{18.8}$ & $\mathbf{14.5}$ & \cellcolor{cpBlue!50}$-4.3$ \\
\bottomrule
\end{tabular}
\caption{Selected dense-subset subgroup results, in percent. Backbone denotes $MLM+T_{id}(16)$, and the head in parentheses is the auxiliary arm compared with it. In the F1 panel, MITweet uses the random split and SemEval the seen targets. In the error panel, BASIL rows report lexical-bias erasure, and A and F denote Against and Favor. Bold marks the better value within each row and underlining the second. \textcolor{cpBlue}{Blue} cells are improvements and \textcolor{cpOrange!80!black}{orange} cells regressions; darker means larger. Appendix~\ref{appendix:results} reports every facet, target, and outlet.}
\label{tab:selected-gains}\label{tab:selected-errors}
\end{table}

\subsection{The error landscape changes with the supervision}
(RQ2) Table~\ref{tab:selected-errors} reports the detailed comparisons discussed here.

On MITweet, gains concentrate by facet. Theme and Tone tie for the lowest random-split reversal rate, but the heads help different ideological dimensions. Under held-out facets, MLM+TLP gives the best overall F1 and reversal rate, so the facet-specific gains do not transfer uniformly.

BASIL separates detection from attribution. Tone gives the highest lexical-bias F1, while joint supervision gives the lowest lexical-bias erasure. Joint supervision lowers erasure at both Fox News and \textit{The New York Times}, but the reduction is more than twice as large at the \textit{Times}. Thus, the same objective changes the outlet comparison as well as the average score.

SemEval also splits by target and direction. On seen targets, joint supervision gives the highest stance F1 and lowers Favor-to-Against reversals, but raises Against-to-Favor reversals. On the held-out Trump target, Tone lowers total reversal relative to MLM+TLP, but MLM+TLP retains the higher stance F1. These results show that the model that best preserves polarity is therefore not always the model that best separates the classes.

These tradeoffs change model choice.
Tone is preferable for BASIL lexical F1, but joint supervision is preferable for reducing missed lexical bias. Researchers should select the checkpoint against the error that would distort the intended comparison, not from F1 alone.

\subsection{Partial label coverage changes the pattern}

(RQ3) The full-corpus runs use the same triplet count and static loss weights, but only the GDELT-matched third receives theme or tone loss. Table~\ref{tab:main-partial} reports changes in the primary scores relative to the dense subset; absolute scores and errors remain in Appendix~\ref{appendix:results}.

Some patterns persist.
Tone remains competitive on BASIL lexical bias. Other rankings move. MITweet changes by about one point or less, with Theme the only arm that improves held-out-facet F1. Seen-target SemEval falls for the tone-bearing arms, while joint supervision loses the least F1 on the held-out target and becomes the strongest full-corpus arm there. 
BASIL informational bias keeps its ranking, but the gap between the backbone and joint supervision narrows from 3.5 to 2.3 points.

\section{Discussion}
\label{sec:discussion}

\paragraph{Metalinguistic benefits are uneven.}
No auxiliary head raises every primary score, and the aggregate differences fall within seed variance (Appendix~\ref{appendix:statistics}). The pattern is sharper below the aggregate: Theme, Tone, and joint supervision lead different MITweet facets; joint supervision improves all five seen SemEval targets over the backbone; and BASIL gains differ by outlet and error type. These subgroup differences suggest that any value of metalinguistic supervision lies in specific facets, outlets, and targets rather than in a uniform score increase.

\paragraph{There is no universal winner.}
Framing and agenda-setting make theme a plausible signal for issue selection, while stance and biased-language work make tone plausible for target evaluation and lexical bias. The results partly follow that motivation: Tone leads BASIL lexical bias and joint supervision leads seen-target stance. 
Theme does not lead BASIL informational bias. That is consistent with a granularity limit: GDELT themes describe an article, whereas BASIL informational bias marks which event details appear in a sentence. The contrastive backbone may also already encode part of the same story-level structure. Weak labels should therefore be tested at the unit where the classifier will be used.

\begin{table}[t]
\centering
\footnotesize
\setlength{\tabcolsep}{2.1pt}
\begin{tabular}{@{}llrrrr@{}}
\toprule
Dataset & F1 & Backbone & $+T_h$ & $+T_o$ & $+T_h+T_o$ \\
\midrule
MITweet & Random split & \cellcolor{cpOrange!15}$\underline{-0.1}$ & \cellcolor{cpBlue!15}$\mathbf{+0.2}$ & \cellcolor{cpOrange!15}$-0.9$ & \cellcolor{cpOrange!15}$-0.5$ \\
 & Held-out facet & \cellcolor{cpOrange!15}$\underline{-0.6}$ & \cellcolor{cpBlue!15}$\mathbf{+0.5}$ & \cellcolor{cpOrange!15}$-0.9$ & \cellcolor{cpOrange!15}$-0.8$ \\
\midrule
BASIL & Info. & \cellcolor{cpBlue!15}$\underline{+0.7}$ & \cellcolor{cpBlue!15}$+0.4$ & \cellcolor{cpOrange!15}$-0.5$ & \cellcolor{cpBlue!30}$\mathbf{+2.0}$ \\
 & Lexical & \cellcolor{cpBlue!50}$\mathbf{+3.4}$ & \cellcolor{cpBlue!15}$\underline{+0.7}$ & \cellcolor{cpBlue!15}$+0.6$ & \cellcolor{cpBlue!15}$+0.3$ \\
\midrule
SemEval & Seen & \cellcolor{cpBlue!30}$\underline{+1.1}$ & \cellcolor{cpBlue!50}$\mathbf{+9.7}$ & \cellcolor{cpOrange!30}$-1.7$ & \cellcolor{cpOrange!30}$-2.5$ \\
 & Held-out & \cellcolor{cpOrange!50}$-8.8$ & \cellcolor{cpOrange!50}$\underline{-3.5}$ & \cellcolor{cpOrange!50}$-4.7$ & \cellcolor{cpOrange!15}$\mathbf{-0.9}$ \\
\bottomrule
\end{tabular}
\caption{Change in primary score under full-corpus partial supervision relative to the dense subset, in percentage points. Backbone denotes $MLM+T_{id}(16)$; SemEval rows split seen and held-out targets. Positive favors the partial regime. \textcolor{cpBlue}{Blue} cells favor the partial regime and \textcolor{cpOrange!80!black}{orange} cells the dense subset; darker means larger. Bold marks the largest improvement and underlining the second largest within each row. Full score and error tables appear in Appendix~\ref{appendix:results}.}
\label{tab:main-partial}
\end{table}

\paragraph{Accuracy and polarity answer different questions.}
The model with the highest F1 and the model with the lowest directional error can differ, as on BASIL lexical bias and seen-target SemEval (Table~\ref{tab:selected-errors}). A study should maximize F1 when overall discrimination is primary, but prefer a lower-reversal or lower-erasure model, subject to acceptable F1, when that error would reverse the substantive conclusion.

\paragraph{Coverage changes the interaction with the backbone.}
To our knowledge, this is the first political-text study to compare the same auxiliary targets under both dense and naturally partial coverage on a fixed triplet-based contrastive backbone. Several rankings change when all articles receive MLM+TLP but only the GDELT-matched third receives auxiliary loss. It captures the real-world scenario where weak metadata is observed for a particular population, and that population determines which gradients modify the shared representation.

\section{Conclusion}
\label{sec:conclusion}

We tested whether semantic-theme and net-tone supervision adds useful information to a political encoder trained with triplet loss and masked language modeling. The answer is conditional.
Auxiliary supervision can lower bias erasure or attribution for particular outlets and improve selected ideological facets or stance targets. It can also worsen another error or fail under context shift.

These results argue against a universal metalinguistic head. They support a design practice: define the political construct, identify its costly errors, test plausible signals against a matched backbone, and audit the result at the outlet, facet, or target level used by the study. The full-corpus runs add a further caution. Correctly masked auxiliary labels can behave differently when they cover only part of the pre-training population.

Two limitations define the next steps. GDELT labels cover only a third of the corpus, so denser or alternative metalinguistic annotations would separate the effect of coverage from that of corpus composition. Each arm also has one RoBERTa-base pre-training run, so repeated runs across model sizes would show whether the facet-, outlet-, and target-level differences replicate.

\section*{Limitations}
\label{sec:limitations}

AllSides is exploratory because its test labels informed the triplet count, epoch, and initial analysis, although its test outlets appear in neither downstream training nor continued pre-training. The configuration selected on AllSides may also suit article-level ideology better than the downstream constructs, which could understate auxiliary gains on MITweet, BASIL, and SemEval.

GDELT themes and net tone are proxies, but they are not perfect representations of the underlying concepts. Themes are not frames or ideological facets. Net tone neither identifies the target of evaluation nor separates positive and negative components. The present experiments show the effects of these objectives; they do not prove that the encoder learned a general theory of theme or tone.

The datasets are English-language and largely U.S.-centered, and several capture earlier political contexts. Political categories, targets, and rhetorical cues vary across countries and periods. The results should not be generalized without new evaluation.

\section*{Ethical Statement}
\label{sec:ethical-statement}

Political classifiers of the kind studied here can support research on ideology, stance, and media bias, but their error costs depend on use. A false positive in retrieval, moderation, or exploratory coding may have a different consequence from the same error in an outlet comparison. Such classifiers can also enable profiling, surveillance, or selective moderation. Researchers should document the intended use, avoid individual-level ideological claims where labels do not support them, and report group-specific failures that could burden particular speakers or communities.

\bibliography{
    bibliography/llm_misc,
    bibliography/ideology_classification,
    bibliography/pluralism,
    bibliography/learning,
    bibliography/metalinguistic_awareness,
    bibliography/aaai_misc
}

\clearpage 
\appendix
\section{Configuration Selection}
\label{appendix:pretraining}

AllSides alone selected the dense-subset configuration and formed the hypothesis. We first trained one-epoch checkpoints with 1, 16, or 32 triplets, then fine-tuned them with the same AllSides source-disjoint procedure. Table~\ref{tab:triplet-selection} shows macro-F1.
\begin{table}[t]
\centering
\setlength{\tabcolsep}{3.0pt}
\resizebox{\linewidth}{!}{%
\begin{tabular}{lcccc}
\toprule
Triplets & TLP+MLM & +Theme & +Tone & +Theme+Tone \\
\midrule
1 & $34.4_{1.4}$ & $33.3_{2.2}$ & $\underline{34.3_{1.8}}$ & $\mathbf{37.1_{4.0}}$ \\
16 & $\mathbf{35.6_{2.9}}$ & $\mathbf{37.3_{3.2}}$ & $\mathbf{34.7_{2.2}}$ & $\underline{34.1_{1.7}}$ \\
32 & $\underline{35.5_{1.8}}$ & $\underline{33.9_{1.8}}$ & $31.3_{1.8}$ & $33.6_{3.2}$ \\
\bottomrule
\end{tabular}%
}
\caption{AllSides test macro-F1 used to select one triplet count on the dense subset after one pre-training epoch. Subscripts give standard deviations across available fine-tuning seeds. Bold marks the best displayed mean in a column; underlining marks the second best.}
\label{tab:triplet-selection}
\end{table}

At 16 triplets, one epoch exceeds two epochs for all four arms (Table~\ref{tab:epoch-selection}). We fixed one epoch for every dense-subset experiment and fixed 16 triplets for the full-corpus runs. MITweet, BASIL, and SemEval use the frozen triplet count and loss weights.

\begin{table}[t]
\centering
\setlength{\tabcolsep}{3.0pt}
\resizebox{\linewidth}{!}{%
\begin{tabular}{lcccc}
\toprule
Epochs & TLP+MLM & +Theme & +Tone & +Theme+Tone \\
\midrule
1 & $\mathbf{35.6_{2.9}}$ & $\mathbf{37.3_{3.2}}$ & $\mathbf{34.7_{2.2}}$ & $\mathbf{34.1_{1.7}}$ \\
2 & $\underline{33.6_{0.7}}$ & $\underline{37.0_{3.5}}$ & $\underline{33.3_{1.4}}$ & $\underline{32.9_{0.3}}$ \\
\bottomrule
\end{tabular}%
}
\caption{AllSides test macro-F1 used to select the dense-subset pre-training epoch at 16 triplets. Subscripts give standard deviations across available fine-tuning seeds. Bold marks the best displayed mean in a column; underlining marks the second best.}
\label{tab:epoch-selection}
\end{table}

\section{Static Loss-Weight Selection}
\label{appendix:weight-selection}

We selected one static weight vector before the final runs. A diagnostic measured each unweighted objective's gradient on the final RoBERTa encoder layer across 12 virtual batches. Each virtual batch averaged eight microbatches of 32 examples, matching the eight-GPU training geometry. The median gradient norms were 34.37 for tone, 4.61 for triplet loss, 0.50 for MLM, and 0.08 for theme prediction.

We fixed MLM at 1.0 and used the conservative rule
\begin{equation}
\tilde{w}_i=\sqrt{\frac{\operatorname{median}\lVert g_{\mathrm{MLM}}\rVert}
{\operatorname{median}\lVert g_i\rVert}}.
\end{equation}
Suggested values were clamped to $[0.2,5.0]$ and floored to a 0.05 grid. This rule reduced, rather than erased, differences in task influence. It suggested 0.33 for triplet loss, 2.48 for theme, and 0.12 for tone. We adopted 0.30, 2.45, and 0.20, with MLM fixed at 1.0, and kept them unchanged across arms and triplet counts.

\begin{table}[t]
\centering
\setlength{\tabcolsep}{3.5pt}
\begin{tabular}{lrrr}
\toprule
Objective & Median $\lVert g_i\rVert$ & Rule & Final \\
\midrule
TLP & 4.61 & 0.33 & 0.30 \\
Theme & 0.08 & 2.48 & 2.45 \\
Tone & 34.37 & 0.12 & 0.20 \\
MLM & 0.50 & 1.00 & 1.00 \\
\bottomrule
\end{tabular}
\caption{Gradient-based static loss-weight selection. The final column gives the weights used in all reported runs.}
\label{tab:loss-weight-selection}
\end{table}

\section{Training Details}
\label{app:training_configurations}

\begin{table}[th]
\centering
\setlength{\tabcolsep}{3.5pt}
\resizebox{\linewidth}{!}{%
\begin{tabular}{ll}
\toprule
Setting & Continued pre-training \\
\midrule
Backbone & RoBERTa-base \\
Optimizer & AdamW \\
Learning rate & $5\times10^{-5}$ \\
Batch size & 32 per GPU \\
Gradient accumulation & 8 steps \\
Gradient clipping & 1.0 \\
Precision & bfloat16 \\
GPUs & 8 $\times$ 96GB RTX PRO 6000 \\
Seed & 42 \\
Dense budget & 1 epoch \\
Triplets per batch & 16 \\
Loss weights & Triplets 0.30; Theme 2.45 \\
& Tone 0.20; MLM 1.00 \\
\bottomrule
\end{tabular}%
}
\caption{Continued pre-training configuration. All four factorial arms share the optimizer, data order, triplet count, and applicable task weights.}
\label{tab:pretraining-config}
\end{table}

The dense-subset runs use one epoch because of the AllSides selection study in Appendix~\ref{appendix:pretraining}. Full-corpus runs keep the selected triplet count and static task weights under a matched schedule within that regime.

\begin{table}[t]
\centering
\setlength{\tabcolsep}{3.5pt}
\begin{tabular}{ll}
\toprule
Setting & Downstream fine-tuning \\
\midrule
Optimizer & AdamW \\
Learning rate & $1\times10^{-5}$ \\
Batch size & 32 \\
Evaluation batch size & 8 \\
Gradient accumulation & 4 steps \\
Warmup ratio & 0.06 \\
Weight decay & 0.01 \\
Precision & float16 \\
GPU & 1 $\times$ 24GB RTX 4090 \\
Seeds & 42, 13, 1, 1234, 6789 \\
Maximum epochs & 10 \\
Selection & Validation macro-F1 \\
\bottomrule
\end{tabular}
\caption{Shared downstream fine-tuning configuration. BASIL uses seed 42 in each of ten folds. AllSides, MITweet, and SemEval use five seeds, except dense-subset SemEval Task A, which uses three (13, 1234, 6789).}
\label{tab:finetuning-config}
\end{table}

Fine-tuning uses early stopping with patience two. Every comparison within a dataset uses the same preprocessing, split, truncation, and selection rule. The archived metric files retain the full training arguments for each run.

\section{Paired Primary-Score Summaries}
\label{appendix:statistics}

Table~\ref{tab:paired-ttests-dense} compares each auxiliary arm with MLM+TLP on the dense subset. We report the paired mean difference and a 95\% paired $t$ interval over matched fine-tuning seeds. For held-out-facet MITweet, each seed is first averaged over the four round-robin folds. All intervals condition on one continued-pre-training checkpoint per arm and omit pre-training variance.

\begin{table*}[t]
\centering
\begin{tabular}{lllrrr}
\toprule
Dataset & Task & Arm & $\Delta$ & 95\% interval & Pairs \\
\midrule
AllSides & Curated source-disjoint & $+T_h$ & $+1.7$ & $[+0.1,\,+3.2]$ & 5 \\
 &  & $+T_o$ & $-0.9$ & $[-2.4,\,+0.7]$ & 5 \\
 &  & $+T_h{+}T_o$ & $-1.5$ & $[-6.1,\,+3.0]$ & 5 \\
\midrule
MITweet & Ideology & $+T_h$ & $-0.0$ & $[-1.1,\,+1.0]$ & 5 \\
 &  & $+T_o$ & $+1.0$ & $[-0.0,\,+1.9]$ & 5 \\
 &  & $+T_h{+}T_o$ & $+0.8$ & $[-0.4,\,+2.0]$ & 5 \\
\cmidrule(lr){2-6}
 & Held-out facets & $+T_h$ & $-0.5$ & $[-1.9,\,+0.9]$ & 5 \\
 &  & $+T_o$ & $-0.2$ & $[-1.6,\,+1.1]$ & 5 \\
 &  & $+T_h{+}T_o$ & $-0.2$ & $[-2.0,\,+1.5]$ & 5 \\
\midrule
BASIL & Informational & $+T_h$ & $-1.5$ & $[-3.6,\,+0.5]$ & 10 \\
 &  & $+T_o$ & $-0.1$ & $[-3.3,\,+3.0]$ & 10 \\
 &  & $+T_h{+}T_o$ & $-3.5$ & $[-6.9,\,-0.2]$ & 10 \\
\cmidrule(lr){2-6}
 & Lexical & $+T_h$ & $+1.7$ & $[-4.1,\,+7.4]$ & 10 \\
 &  & $+T_o$ & $+3.2$ & $[-2.1,\,+8.5]$ & 10 \\
 &  & $+T_h{+}T_o$ & $+2.4$ & $[-3.2,\,+7.9]$ & 10 \\
\midrule
SemEval & Task A & $+T_h$ & $-11.9$ & $[-61.9,\,+38.1]$ & 3 \\
 &  & $+T_o$ & $+0.8$ & $[-0.5,\,+2.1]$ & 3 \\
 &  & $+T_h{+}T_o$ & $+1.2$ & $[-0.4,\,+2.7]$ & 3 \\
\cmidrule(lr){2-6}
 & Task B & $+T_h$ & $-3.5$ & $[-16.3,\,+9.2]$ & 5 \\
 &  & $+T_o$ & $-3.3$ & $[-11.9,\,+5.3]$ & 5 \\
 &  & $+T_h{+}T_o$ & $-3.7$ & $[-15.2,\,+7.8]$ & 5 \\
\bottomrule
\end{tabular}
\caption{Dense-subset paired differences in the primary task score, in percentage points, for each auxiliary arm against $MLM{+}T_{id}(16)$. Intervals are paired $t$ intervals over matched fine-tuning seeds. Pairs are five matched seeds for AllSides, MITweet, and SemEval Task B, three for SemEval Task A, and ten matched folds for BASIL.}
\label{tab:paired-ttests-dense}
\end{table*}

\section{Complete Dense- and Partial-Supervision Results}
\label{appendix:results}

This appendix gives the absolute task and error results for both supervision regimes. The dense-subset block reports the controlled setting in which every article has valid theme and tone targets. The full-corpus block reports the naturally partial setting in which every eligible article receives MLM and TLP, while only the GDELT-matched third receives auxiliary loss. Bold marks the best displayed value and underlining the second-best distinct value.

Tables~\ref{tab:app-allsides-dense}--\ref{tab:app-semeval-dense} report the dense subset. Tables~\ref{tab:app-allsides-full}--\ref{tab:app-semeval-full} report the full corpus with partial auxiliary supervision.

\begin{table*}[t]
\centering
\setlength{\tabcolsep}{3.2pt}
\resizebox{\linewidth}{!}{%
\begin{tabular}{lcccccccc}
\toprule
 & \multicolumn{4}{c}{Class F1} & \multicolumn{4}{c}{Error geometry} \\
\cmidrule(lr){2-5} \cmidrule(lr){6-9}
Model & Macro-F1 $\uparrow$ & Left $\uparrow$ & Center $\uparrow$ & Right $\uparrow$ & Reversal $\downarrow$ & L$\rightarrow$R $\downarrow$ & R$\rightarrow$L $\downarrow$ & Endpoint acc. $\uparrow$ \\
\midrule
RoBERTa & $34.4_{ 1.8}$ & $43.5_{ 4.8}$ & $\mathbf{22.5_{ 3.6}}$ & $37.3_{ 2.0}$ & $42.7_{ 1.6}$ & $48.0_{ 2.2}$ & $\underline{28.2_{ 5.4}}$ & $\mathbf{42.2_{ 3.3}}$ \\
$MLM+T_{id}(16)$ & $\underline{35.6_{ 2.9}}$ & $41.3_{ 3.1}$ & $17.8_{ 5.8}$ & $\mathbf{47.8_{ 4.1}}$ & $\mathbf{21.3_{ 5.1}}$ & $\underline{19.9_{ 6.5}}$ & $\mathbf{25.1_{ 8.3}}$ & $39.2_{ 2.8}$ \\
$+T_h$ & $\mathbf{37.3_{ 3.2}}$ & $\underline{43.8_{ 2.7}}$ & $\underline{20.6_{ 4.4}}$ & $\underline{47.5_{ 4.7}}$ & $\underline{21.9_{ 7.1}}$ & $\mathbf{19.5_{ 9.3}}$ & $28.8_{ 5.7}$ & $40.7_{ 1.7}$ \\
$+T_o$ & $34.7_{ 2.2}$ & $\mathbf{44.5_{ 3.5}}$ & $17.7_{ 4.3}$ & $42.0_{ 2.1}$ & $28.8_{ 6.8}$ & $27.2_{ 11.4}$ & $33.1_{ 9.9}$ & $\underline{42.0_{ 2.5}}$ \\
$+T_h+T_o$ & $34.1_{ 1.7}$ & $40.4_{ 3.7}$ & $16.1_{ 4.5}$ & $45.8_{ 6.6}$ & $26.1_{ 11.3}$ & $23.3_{ 17.1}$ & $33.6_{ 8.7}$ & $39.1_{ 2.3}$ \\
\bottomrule
\end{tabular}%
}
\caption{AllSides article ideology in the dense subset: per-class F1 and error geometry behind the macro-F1 and reversal rate summarized in Table~\ref{tab:main-dense}. Reversal and its two directions are measured over gold Left/Right articles; endpoint accuracy shows whether a model lowers reversals by predicting Center more often. Subscripts give standard deviations across five fine-tuning seeds.}
\label{tab:app-allsides-dense}
\end{table*}

\begin{sidewaystable*}[p]
\centering
\setlength{\tabcolsep}{2.0pt}
\begin{tabular}{llcccccccccccccc}
\toprule
 &  & \multicolumn{2}{c}{Overall} & \multicolumn{12}{c}{Ideology F1 by facet} \\
\cmidrule(lr){3-4} \cmidrule(lr){5-16}
Split & Model & \makecell{Macro-F1\\$\uparrow$} & \makecell{Reversal\\$\downarrow$} & PoR $\uparrow$ & SS $\uparrow$ & EO $\uparrow$ & EE $\uparrow$ & EP $\uparrow$ & CSR $\uparrow$ & CV $\uparrow$ & DS $\uparrow$ & MF $\uparrow$ & SD $\uparrow$ & JO $\uparrow$ & PeR $\uparrow$ \\
\midrule
Random & RoBERTa & $50.7_{ 0.7}$ & $33.1_{ 2.1}$ & $35.1_{ 9.4}$ & $\underline{33.8_{ 3.1}}$ & $43.6_{ 3.1}$ & $\mathbf{49.4_{ 1.7}}$ & $41.9_{ 1.9}$ & $17.6_{ 6.4}$ & $\mathbf{43.7_{ 7.5}}$ & $42.0_{ 1.3}$ & $45.4_{ 3.5}$ & $\underline{41.6_{ 1.8}}$ & $30.9_{ 2.8}$ & $25.1_{ 2.4}$ \\
 & $MLM+T_{id}(16)$ & $50.1_{ 0.5}$ & $\underline{32.7_{ 1.0}}$ & $\underline{35.7_{ 3.9}}$ & $30.9_{ 4.7}$ & $\mathbf{46.7_{ 7.5}}$ & $44.8_{ 2.6}$ & $\underline{43.6_{ 3.7}}$ & $18.5_{ 8.4}$ & $41.5_{ 8.2}$ & $42.8_{ 0.9}$ & $47.5_{ 4.5}$ & $39.9_{ 1.5}$ & $\underline{31.4_{ 2.6}}$ & $24.8_{ 5.5}$ \\
 & $+T_h$ & $50.1_{ 0.7}$ & $\mathbf{32.4_{ 1.3}}$ & $\mathbf{36.7_{ 7.6}}$ & $32.6_{ 3.7}$ & $44.4_{ 4.1}$ & $46.6_{ 1.1}$ & $43.4_{ 1.3}$ & $15.6_{ 8.5}$ & $\underline{41.8_{ 14.0}}$ & $43.4_{ 2.2}$ & $48.3_{ 4.6}$ & $\mathbf{41.9_{ 1.8}}$ & $\mathbf{32.7_{ 1.7}}$ & $22.0_{ 3.5}$ \\
 & $+T_o$ & $\mathbf{51.1_{ 0.5}}$ & $\mathbf{32.4_{ 0.8}}$ & $34.4_{ 6.9}$ & $\mathbf{34.3_{ 3.2}}$ & $\underline{45.0_{ 2.7}}$ & $\underline{47.8_{ 3.4}}$ & $\mathbf{44.6_{ 3.0}}$ & $\underline{19.8_{ 8.4}}$ & $32.6_{ 8.7}$ & $\mathbf{45.2_{ 1.6}}$ & $\underline{51.8_{ 3.6}}$ & $41.0_{ 1.7}$ & $31.0_{ 1.5}$ & $\mathbf{26.9_{ 2.3}}$ \\
 & $+T_h+T_o$ & $\underline{50.9_{ 0.7}}$ & $32.8_{ 1.5}$ & $32.1_{ 10.4}$ & $33.0_{ 4.8}$ & $\mathbf{46.7_{ 2.0}}$ & $\mathbf{49.4_{ 1.9}}$ & $41.8_{ 1.5}$ & $\mathbf{20.1_{ 8.2}}$ & $37.5_{ 11.3}$ & $\underline{44.8_{ 1.1}}$ & $\mathbf{51.9_{ 4.7}}$ & $40.6_{ 2.1}$ & $30.9_{ 2.5}$ & $\underline{25.8_{ 4.2}}$ \\
\midrule
Held-out & RoBERTa & $27.5_{ 7.9}$ & $62.3_{ 16.0}$ & $26.6_{ 3.7}$ & $27.7_{ 5.0}$ & $39.1_{ 3.5}$ & $\underline{47.1_{ 3.0}}$ & $15.6_{ 2.9}$ & $\underline{23.1_{ 8.6}}$ & $29.0_{ 9.6}$ & $\mathbf{32.1_{ 2.4}}$ & $19.5_{ 0.9}$ & $38.3_{ 1.6}$ & $21.4_{ 2.1}$ & $\underline{8.9_{ 1.6}}$ \\
facet & $MLM+T_{id}(16)$ & $\mathbf{28.1_{ 7.7}}$ & $\mathbf{60.6_{ 16.4}}$ & $\mathbf{29.8_{ 5.5}}$ & $\mathbf{31.8_{ 2.2}}$ & $\mathbf{41.5_{ 2.2}}$ & $46.7_{ 3.9}$ & $\mathbf{19.3_{ 2.6}}$ & $13.6_{ 7.1}$ & $\mathbf{44.1_{ 8.1}}$ & $30.2_{ 2.2}$ & $19.5_{ 1.8}$ & $38.0_{ 2.3}$ & $\mathbf{23.5_{ 2.6}}$ & $8.3_{ 0.6}$ \\
 & $+T_h$ & $27.6_{ 7.5}$ & $\underline{61.8_{ 16.1}}$ & $25.8_{ 4.7}$ & $30.3_{ 4.0}$ & $\underline{41.2_{ 1.0}}$ & $\mathbf{48.4_{ 3.8}}$ & $17.4_{ 1.9}$ & $9.8_{ 3.5}$ & $\underline{38.0_{ 8.1}}$ & $30.6_{ 2.7}$ & $\underline{20.8_{ 1.1}}$ & $37.6_{ 4.1}$ & $\underline{22.3_{ 2.0}}$ & $8.4_{ 0.7}$ \\
 & $+T_o$ & $\underline{27.9_{ 7.9}}$ & $\underline{61.8_{ 17.1}}$ & $\underline{28.2_{ 8.9}}$ & $\underline{30.7_{ 3.4}}$ & $39.5_{ 3.3}$ & $45.5_{ 2.6}$ & $16.7_{ 3.9}$ & $19.5_{ 10.2}$ & $31.9_{ 6.0}$ & $\underline{31.5_{ 3.3}}$ & $19.7_{ 1.4}$ & $\underline{39.2_{ 1.8}}$ & $21.0_{ 2.1}$ & $\mathbf{10.2_{ 1.0}}$ \\
 & $+T_h+T_o$ & $\underline{27.9_{ 7.8}}$ & $\underline{61.8_{ 16.6}}$ & $23.1_{ 10.0}$ & $24.2_{ 5.7}$ & $38.8_{ 2.6}$ & $45.3_{ 2.4}$ & $\underline{17.6_{ 3.1}}$ & $\mathbf{24.0_{ 8.2}}$ & $32.6_{ 6.5}$ & $31.3_{ 2.5}$ & $\mathbf{21.0_{ 2.1}}$ & $\mathbf{39.6_{ 1.8}}$ & $22.2_{ 1.9}$ & $8.2_{ 1.3}$ \\
\bottomrule
\end{tabular}
\caption{MITweet ideology in the dense subset, by split and facet. Overall columns give the macro-F1 and reversal rate summarized in Table~\ref{tab:main-dense}; facet columns give per-facet F1. Under the held-out-facet split each fold evaluates only the facets it withheld, so a facet column averages the folds that held it out. Facet abbreviations follow the dataset. Subscripts give standard deviations across five fine-tuning seeds, except the held-out-facet overall columns, which pool all 20 fold--seed runs.}
\label{tab:app-mitweet-dense}
\end{sidewaystable*}

\begin{sidewaystable*}[p]
\centering
\setlength{\tabcolsep}{3.0pt}
\begin{tabular}{llcccccccccccc}
\toprule
 &  & \multicolumn{3}{c}{Overall} & \multicolumn{3}{c}{Fox} & \multicolumn{3}{c}{HuffPost} & \multicolumn{3}{c}{NYT} \\
\cmidrule(lr){3-5} \cmidrule(lr){6-8} \cmidrule(lr){9-11} \cmidrule(lr){12-14}
Bias & Model & F1 $\uparrow$ & Erase $\downarrow$ & Attrib. $\downarrow$ & F1 $\uparrow$ & Erase $\downarrow$ & Attrib. $\downarrow$ & F1 $\uparrow$ & Erase $\downarrow$ & Attrib. $\downarrow$ & F1 $\uparrow$ & Erase $\downarrow$ & Attrib. $\downarrow$ \\
\midrule
Informational & RoBERTa & $45.0_{ 4.5}$ & $\underline{38.9_{ 6.2}}$ & $20.2_{ 5.6}$ & $\underline{47.7_{ 8.1}}$ & $\mathbf{34.7_{ 10.5}}$ & $21.0_{ 6.0}$ & $\underline{46.7_{ 8.5}}$ & $\mathbf{36.5_{ 11.3}}$ & $20.2_{ 7.2}$ & $\underline{39.8_{ 10.1}}$ & $45.5_{ 14.4}$ & $\underline{19.6_{ 5.4}}$ \\
 & $MLM+T_{id}(16)$ & $\mathbf{45.6_{ 3.3}}$ & $\mathbf{38.0_{ 7.4}}$ & $19.7_{ 3.3}$ & $47.3_{ 10.1}$ & $\underline{35.8_{ 14.3}}$ & $20.4_{ 5.7}$ & $\mathbf{47.4_{ 8.4}}$ & $\underline{36.9_{ 11.2}}$ & $\underline{18.7_{ 4.8}}$ & $\mathbf{41.5_{ 7.0}}$ & $\mathbf{42.6_{ 12.0}}$ & $19.7_{ 4.7}$ \\
 & $+T_h$ & $44.1_{ 3.5}$ & $41.0_{ 8.2}$ & $\underline{19.5_{ 4.3}}$ & $\underline{47.7_{ 12.0}}$ & $36.6_{ 16.9}$ & $\underline{19.2_{ 6.4}}$ & $43.3_{ 6.9}$ & $43.0_{ 9.2}$ & $19.6_{ 6.3}$ & $39.7_{ 9.4}$ & $\underline{45.4_{ 14.0}}$ & $19.9_{ 4.5}$ \\
 & $+T_o$ & $\underline{45.5_{ 4.2}}$ & $41.0_{ 6.6}$ & $\mathbf{18.2_{ 3.7}}$ & $\mathbf{49.0_{ 9.4}}$ & $35.9_{ 15.7}$ & $\mathbf{18.1_{ 4.8}}$ & $44.7_{ 8.0}$ & $43.5_{ 8.2}$ & $\mathbf{17.8_{ 5.4}}$ & $\mathbf{41.5_{ 8.8}}$ & $\underline{45.4_{ 8.3}}$ & $\mathbf{18.4_{ 4.6}}$ \\
 & $+T_h+T_o$ & $42.1_{ 5.4}$ & $43.3_{ 7.9}$ & $20.4_{ 3.7}$ & $44.0_{ 11.5}$ & $40.9_{ 18.3}$ & $20.3_{ 4.2}$ & $44.1_{ 10.6}$ & $41.9_{ 10.6}$ & $20.0_{ 6.5}$ & $37.5_{ 6.4}$ & $48.4_{ 8.2}$ & $20.7_{ 4.1}$ \\
\midrule
Lexical & RoBERTa & $28.4_{ 5.6}$ & $46.8_{ 14.7}$ & $12.8_{ 3.3}$ & $20.5_{ 10.3}$ & $57.2_{ 26.3}$ & $\underline{13.8_{ 4.1}}$ & $32.5_{ 15.6}$ & $43.8_{ 30.7}$ & $\underline{11.8_{ 3.6}}$ & $\mathbf{28.6_{ 8.0}}$ & $\underline{39.2_{ 13.2}}$ & $12.9_{ 4.8}$ \\
 & $MLM+T_{id}(16)$ & $26.8_{ 8.5}$ & $51.4_{ 18.9}$ & $\underline{12.2_{ 4.6}}$ & $23.4_{ 10.9}$ & $51.6_{ 25.3}$ & $14.1_{ 5.1}$ & $28.4_{ 19.3}$ & $52.4_{ 30.3}$ & $\mathbf{11.2_{ 5.0}}$ & $25.9_{ 11.7}$ & $52.7_{ 12.5}$ & $\mathbf{11.6_{ 5.4}}$ \\
 & $+T_h$ & $28.5_{ 7.1}$ & $50.0_{ 15.5}$ & $\mathbf{12.1_{ 6.2}}$ & $\mathbf{24.4_{ 10.5}}$ & $51.6_{ 23.1}$ & $\mathbf{12.7_{ 4.7}}$ & $28.2_{ 15.9}$ & $55.8_{ 22.2}$ & $\underline{11.8_{ 7.2}}$ & $\underline{28.5_{ 10.8}}$ & $44.2_{ 21.3}$ & $\underline{12.0_{ 7.0}}$ \\
 & $+T_o$ & $\mathbf{30.0_{ 5.6}}$ & $\underline{43.6_{ 9.6}}$ & $13.4_{ 4.7}$ & $22.8_{ 10.0}$ & $\underline{51.2_{ 21.9}}$ & $14.8_{ 5.6}$ & $\mathbf{35.4_{ 13.5}}$ & $\underline{39.6_{ 20.7}}$ & $12.5_{ 5.5}$ & $27.3_{ 12.1}$ & $42.7_{ 21.9}$ & $12.8_{ 4.3}$ \\
 & $+T_h+T_o$ & $\underline{29.1_{ 4.0}}$ & $\mathbf{39.1_{ 8.6}}$ & $15.4_{ 4.7}$ & $\underline{24.2_{ 8.2}}$ & $\mathbf{45.4_{ 19.9}}$ & $15.5_{ 5.8}$ & $\underline{34.8_{ 12.6}}$ & $\mathbf{37.1_{ 15.8}}$ & $14.7_{ 5.1}$ & $25.9_{ 8.1}$ & $\mathbf{36.9_{ 17.2}}$ & $15.9_{ 5.0}$ \\
\bottomrule
\end{tabular}
\caption{BASIL sentence-level bias in the dense subset, by annotation type and outlet. F1 is positive-class F1. Erasure is the false-negative rate, which misses an annotated instance of bias; attribution is the false-positive rate, which marks an unannotated sentence as biased. Overall columns give the values summarized in Table~\ref{tab:main-dense}. Subscripts give standard deviations across ten folds, each with one fine-tuning seed.}
\label{tab:app-basil-dense}
\end{sidewaystable*}

\begin{table*}[t]
\centering
\setlength{\tabcolsep}{2.6pt}
\textbf{Task A: seen targets}\par\smallskip
\resizebox{\linewidth}{!}{%
\begin{tabular}{lccccccccc}
\toprule
Model & Stance F1 $\uparrow$ & Reversal $\downarrow$ & A$\rightarrow$F $\downarrow$ & F$\rightarrow$A $\downarrow$ & Atheism $\uparrow$ & Climate $\uparrow$ & Feminism $\uparrow$ & Hillary $\uparrow$ & Abortion $\uparrow$ \\
\midrule
RoBERTa & $70.8_{ 0.7}$ & $18.6_{ 0.8}$ & $17.5_{ 2.9}$ & $\mathbf{21.3_{ 4.2}}$ & $\mathbf{62.2_{ 10.7}}$ & $44.4_{ 0.3}$ & $57.4_{ 2.0}$ & $\mathbf{73.7_{ 1.4}}$ & $56.6_{ 4.4}$ \\
$MLM+T_{id}(16)$ & $70.3_{ 1.5}$ & $\mathbf{17.7_{ 1.1}}$ & $\underline{12.8_{ 1.9}}$ & $29.3_{ 8.2}$ & $56.9_{ 7.6}$ & $44.7_{ 0.5}$ & $\underline{58.0_{ 0.4}}$ & $68.3_{ 11.1}$ & $53.8_{ 5.6}$ \\
$+T_h$ & $58.4_{ 19.1}$ & $22.1_{ 6.8}$ & $\mathbf{8.2_{ 7.1}}$ & $54.8_{ 39.1}$ & $50.5_{ 10.3}$ & $31.9_{ 22.4}$ & $49.3_{ 9.2}$ & $59.3_{ 19.6}$ & $45.9_{ 9.0}$ \\
$+T_o$ & $\underline{71.1_{ 1.1}}$ & $18.7_{ 0.7}$ & $16.2_{ 4.0}$ & $24.6_{ 7.7}$ & $\underline{62.1_{ 11.3}}$ & $\underline{45.0_{ 0.5}}$ & $57.7_{ 1.7}$ & $69.2_{ 3.5}$ & $\underline{58.9_{ 6.8}}$ \\
$+T_h+T_o$ & $\mathbf{71.5_{ 1.7}}$ & $\underline{17.8_{ 0.8}}$ & $15.2_{ 4.2}$ & $\underline{24.1_{ 7.6}}$ & $61.0_{ 10.8}$ & $\mathbf{45.2_{ 0.5}}$ & $\mathbf{58.4_{ 1.2}}$ & $\underline{71.4_{ 6.4}}$ & $\mathbf{59.0_{ 6.7}}$ \\
\bottomrule
\end{tabular}%
}
\par\medskip
\textbf{Task B: held-out target (Trump)}\par\smallskip
\begin{tabular}{lcccc}
\toprule
Model & Stance F1 $\uparrow$ & Reversal $\downarrow$ & A$\rightarrow$F $\downarrow$ & F$\rightarrow$A $\downarrow$ \\
\midrule
RoBERTa & $19.9_{ 6.9}$ & $\mathbf{9.8_{ 3.7}}$ & $\mathbf{0.5_{ 0.6}}$ & $\mathbf{28.6_{ 11.2}}$ \\
$MLM+T_{id}(16)$ & $\mathbf{34.5_{ 7.4}}$ & $18.8_{ 5.0}$ & $2.8_{ 2.0}$ & $51.2_{ 16.7}$ \\
$+T_h$ & $31.0_{ 8.8}$ & $24.6_{ 9.9}$ & $\underline{1.4_{ 1.8}}$ & $71.4_{ 30.7}$ \\
$+T_o$ & $\underline{31.2_{ 1.4}}$ & $\underline{14.5_{ 2.3}}$ & $1.9_{ 1.0}$ & $\underline{39.9_{ 8.0}}$ \\
$+T_h+T_o$ & $30.8_{ 8.2}$ & $15.3_{ 3.2}$ & $1.7_{ 1.0}$ & $42.8_{ 8.7}$ \\
\bottomrule
\end{tabular}
\caption{SemEval target-conditioned stance in the dense subset. Stance F1 is the mean of Against and Favor F1, as in Table~\ref{tab:main-dense}. Reversal is endpoint-to-endpoint error, split into its two directions, where A and F denote Against and Favor. Task A's remaining columns give the same stance F1 computed over each target's test examples. The panels are listed separately because Task A has five seen targets while Task B has one held-out target, whose per-target score is the panel score. Subscripts give standard deviations across three fine-tuning seeds for Task A and five for Task B.}
\label{tab:app-semeval-dense}
\end{table*}

\begin{table*}[t]
\centering
\setlength{\tabcolsep}{3.2pt}
\resizebox{\linewidth}{!}{%
\begin{tabular}{lcccccccc}
\toprule
 & \multicolumn{4}{c}{Class F1} & \multicolumn{4}{c}{Error geometry} \\
\cmidrule(lr){2-5} \cmidrule(lr){6-9}
Model & Macro-F1 $\uparrow$ & Left $\uparrow$ & Center $\uparrow$ & Right $\uparrow$ & Reversal $\downarrow$ & L$\rightarrow$R $\downarrow$ & R$\rightarrow$L $\downarrow$ & Endpoint acc. $\uparrow$ \\
\midrule
RoBERTa & $34.9_{ 5.9}$ & $46.3_{ 12.4}$ & $\mathbf{21.1_{ 6.2}}$ & $37.2_{ 4.8}$ & $36.9_{ 7.1}$ & $37.8_{ 9.5}$ & $34.4_{ 4.7}$ & $44.0_{ 11.7}$ \\
$MLM+T_{id}(16)$ & $36.2_{ 1.2}$ & $39.1_{ 2.8}$ & $13.0_{ 1.3}$ & $\mathbf{56.5_{ 2.1}}$ & $\mathbf{17.0_{ 2.7}}$ & $12.8_{ 3.6}$ & $\mathbf{28.6_{ 4.0}}$ & $39.7_{ 2.6}$ \\
$+T_h$ & $\mathbf{38.6_{ 2.8}}$ & $\mathbf{48.0_{ 5.7}}$ & $12.4_{ 2.9}$ & $\underline{55.6_{ 3.8}}$ & $\underline{17.2_{ 3.6}}$ & $\mathbf{11.0_{ 5.3}}$ & $34.2_{ 3.8}$ & $\mathbf{46.7_{ 5.1}}$ \\
$+T_o$ & $35.9_{ 1.6}$ & $42.2_{ 3.6}$ & $16.2_{ 4.7}$ & $49.3_{ 2.7}$ & $17.9_{ 3.9}$ & $\underline{12.1_{ 4.7}}$ & $33.8_{ 5.5}$ & $39.3_{ 3.2}$ \\
$+T_h+T_o$ & $\underline{38.5_{ 4.9}}$ & $\underline{46.8_{ 7.9}}$ & $\underline{17.1_{ 4.1}}$ & $51.7_{ 4.3}$ & $18.8_{ 4.7}$ & $14.5_{ 4.2}$ & $\underline{30.7_{ 7.8}}$ & $\underline{44.4_{ 7.2}}$ \\
\bottomrule
\end{tabular}%
}
\caption{AllSides article ideology under full-corpus partial auxiliary supervision. Reversal and its two directions are measured over gold Left/Right articles; endpoint accuracy checks whether a lower reversal rate comes from predicting Center more often. Subscripts give standard deviations across five fine-tuning seeds.}
\label{tab:app-allsides-full}
\end{table*}

\begin{sidewaystable*}[p]
\centering
\setlength{\tabcolsep}{2.0pt}
\textbf{Random split}\par\smallskip
\begin{tabular}{lcccccccccccccc}
\toprule
Model & \makecell{Macro-F1\\$\uparrow$} & \makecell{Reversal\\$\downarrow$} & PoR $\uparrow$ & SS $\uparrow$ & EO $\uparrow$ & EE $\uparrow$ & EP $\uparrow$ & CSR $\uparrow$ & CV $\uparrow$ & DS $\uparrow$ & MF $\uparrow$ & SD $\uparrow$ & JO $\uparrow$ & PeR $\uparrow$ \\
\midrule
RoBERTa & $\mathbf{50.6_{ 0.7}}$ & $33.4_{ 1.7}$ & $\underline{36.6_{ 7.8}}$ & $33.8_{ 4.2}$ & $43.5_{ 4.2}$ & $\mathbf{50.8_{ 2.0}}$ & $40.8_{ 2.0}$ & $\underline{14.8_{ 8.9}}$ & $\mathbf{42.9_{ 9.1}}$ & $42.0_{ 1.1}$ & $45.1_{ 1.8}$ & $40.5_{ 2.1}$ & $30.8_{ 1.7}$ & $\underline{25.0_{ 1.4}}$ \\
$MLM+T_{id}(16)$ & $50.0_{ 0.5}$ & $\mathbf{32.3_{ 0.8}}$ & $32.9_{ 14.2}$ & $\mathbf{37.5_{ 5.3}}$ & $44.7_{ 3.7}$ & $46.0_{ 4.4}$ & $42.6_{ 2.9}$ & $9.3_{ 5.9}$ & $\underline{38.8_{ 12.2}}$ & $41.6_{ 1.7}$ & $48.3_{ 4.5}$ & $40.3_{ 2.3}$ & $\mathbf{32.8_{ 2.8}}$ & $23.8_{ 3.5}$ \\
$+T_h$ & $50.3_{ 0.5}$ & $\underline{32.6_{ 1.0}}$ & $\mathbf{37.4_{ 15.6}}$ & $\underline{37.2_{ 1.1}}$ & $43.6_{ 3.1}$ & $45.1_{ 2.4}$ & $\mathbf{43.5_{ 3.0}}$ & $9.5_{ 3.5}$ & $34.5_{ 9.8}$ & $\mathbf{43.3_{ 1.4}}$ & $45.5_{ 2.1}$ & $\mathbf{41.3_{ 1.8}}$ & $\underline{32.2_{ 2.5}}$ & $24.7_{ 2.0}$ \\
$+T_o$ & $50.2_{ 0.4}$ & $32.9_{ 1.1}$ & $35.2_{ 3.8}$ & $32.8_{ 1.7}$ & $\mathbf{47.5_{ 2.2}}$ & $44.7_{ 1.1}$ & $\underline{42.7_{ 1.6}}$ & $13.2_{ 6.0}$ & $27.7_{ 3.0}$ & $\underline{43.2_{ 1.9}}$ & $\underline{51.7_{ 3.4}}$ & $\mathbf{41.3_{ 0.8}}$ & $32.0_{ 0.9}$ & $\mathbf{25.2_{ 1.6}}$ \\
$+T_h+T_o$ & $\underline{50.4_{ 0.3}}$ & $32.8_{ 1.3}$ & $36.3_{ 3.1}$ & $34.6_{ 4.3}$ & $\underline{46.9_{ 4.4}}$ & $\underline{47.2_{ 2.5}}$ & $41.2_{ 1.9}$ & $\mathbf{18.7_{ 6.3}}$ & $31.6_{ 6.9}$ & $42.9_{ 0.9}$ & $\mathbf{52.5_{ 5.0}}$ & $\underline{40.8_{ 1.6}}$ & $31.2_{ 1.8}$ & $\mathbf{25.2_{ 1.3}}$ \\
\bottomrule
\end{tabular}
\par\medskip
\textbf{Round-robin held-out facets}\par\smallskip
\begin{tabular}{lcccccc}
\toprule
Model & Macro-F1 $\uparrow$ & Reversal $\downarrow$ & Fold 0 $\uparrow$ & Fold 1 $\uparrow$ & Fold 2 $\uparrow$ & Fold 3 $\uparrow$ \\
\midrule
RoBERTa & $26.8_{ 7.5}$ & $62.9_{ 16.0}$ & $32.1_{ 2.3}$ & $21.0_{ 1.1}$ & $35.5_{ 1.6}$ & $18.5_{ 1.6}$ \\
$MLM+T_{id}(16)$ & $\underline{27.5_{ 7.6}}$ & $\mathbf{61.4_{ 16.6}}$ & $\underline{32.4_{ 2.3}}$ & $21.2_{ 1.2}$ & $36.7_{ 2.2}$ & $\mathbf{19.8_{ 0.9}}$ \\
$+T_h$ & $\mathbf{28.1_{ 8.1}}$ & $\underline{61.6_{ 16.2}}$ & $\underline{32.4_{ 0.5}}$ & $\mathbf{22.5_{ 1.0}}$ & $\mathbf{38.5_{ 2.1}}$ & $\underline{18.9_{ 1.2}}$ \\
$+T_o$ & $26.9_{ 8.6}$ & $62.5_{ 16.1}$ & $\mathbf{32.6_{ 0.6}}$ & $20.7_{ 1.4}$ & $37.4_{ 1.2}$ & $17.0_{ 1.3}$ \\
$+T_h+T_o$ & $27.0_{ 8.4}$ & $62.1_{ 17.0}$ & $31.0_{ 1.7}$ & $\underline{21.7_{ 1.2}}$ & $\underline{38.1_{ 2.1}}$ & $17.2_{ 0.4}$ \\
\bottomrule
\end{tabular}
\caption{MITweet ideology under full-corpus partial auxiliary supervision. The random-split panel reports overall and facet-level F1. The held-out panel reports the mean over the four round-robin folds and each fold-level macro-F1; each fold withholds a disjoint subset of facets. Facet abbreviations follow the dataset. Subscripts give standard deviations across five fine-tuning seeds, except the held-out overall columns, which pool all 20 fold--seed runs.}
\label{tab:app-mitweet-full}
\end{sidewaystable*}

\begin{sidewaystable*}[p]
\centering
\setlength{\tabcolsep}{3.0pt}
\begin{tabular}{llcccccccccccc}
\toprule
 &  & \multicolumn{3}{c}{Overall} & \multicolumn{3}{c}{Fox} & \multicolumn{3}{c}{HuffPost} & \multicolumn{3}{c}{NYT} \\
\cmidrule(lr){3-5} \cmidrule(lr){6-8} \cmidrule(lr){9-11} \cmidrule(lr){12-14}
Bias & Model & F1 $\uparrow$ & Erase $\downarrow$ & Attrib. $\downarrow$ & F1 $\uparrow$ & Erase $\downarrow$ & Attrib. $\downarrow$ & F1 $\uparrow$ & Erase $\downarrow$ & Attrib. $\downarrow$ & F1 $\uparrow$ & Erase $\downarrow$ & Attrib. $\downarrow$ \\
\midrule
Informational & RoBERTa & $43.5_{ 4.4}$ & $41.8_{ 6.1}$ & $\underline{19.9_{ 5.5}}$ & $46.3_{ 8.2}$ & $37.9_{ 12.9}$ & $20.4_{ 6.1}$ & $45.1_{ 8.4}$ & $40.7_{ 11.2}$ & $\mathbf{19.4_{ 7.4}}$ & $38.6_{ 10.5}$ & $47.6_{ 12.1}$ & $\underline{19.9_{ 5.7}}$ \\
 & $MLM+T_{id}(16)$ & $\mathbf{46.4_{ 3.9}}$ & $36.7_{ 8.7}$ & $\mathbf{19.7_{ 4.6}}$ & $\mathbf{50.9_{ 6.3}}$ & $\underline{31.4_{ 9.0}}$ & $\underline{19.4_{ 5.3}}$ & $\underline{45.9_{ 9.9}}$ & $\underline{36.0_{ 16.0}}$ & $20.5_{ 4.1}$ & $\mathbf{41.5_{ 8.5}}$ & $43.2_{ 12.1}$ & $\mathbf{19.6_{ 6.0}}$ \\
 & $+T_h$ & $44.5_{ 3.1}$ & $39.7_{ 5.9}$ & $20.1_{ 4.3}$ & $46.4_{ 9.9}$ & $38.5_{ 16.1}$ & $\mathbf{19.2_{ 5.4}}$ & $45.1_{ 8.9}$ & $39.7_{ 11.0}$ & $\underline{19.6_{ 4.6}}$ & $\underline{40.4_{ 9.6}}$ & $42.7_{ 11.4}$ & $21.2_{ 5.1}$ \\
 & $+T_o$ & $\underline{45.0_{ 4.0}}$ & $\mathbf{34.6_{ 7.6}}$ & $22.8_{ 5.2}$ & $\underline{49.2_{ 8.2}}$ & $\mathbf{31.3_{ 14.6}}$ & $21.2_{ 7.0}$ & $\mathbf{46.3_{ 7.8}}$ & $\mathbf{32.1_{ 9.9}}$ & $23.2_{ 6.4}$ & $39.3_{ 6.7}$ & $\underline{40.8_{ 8.7}}$ & $23.8_{ 5.4}$ \\
 & $+T_h+T_o$ & $44.1_{ 2.5}$ & $\underline{36.4_{ 9.3}}$ & $22.4_{ 5.5}$ & $46.8_{ 11.3}$ & $35.3_{ 19.7}$ & $20.5_{ 6.1}$ & $45.1_{ 7.9}$ & $36.4_{ 10.4}$ & $21.9_{ 6.2}$ & $39.3_{ 8.8}$ & $\mathbf{39.3_{ 15.5}}$ & $24.5_{ 6.7}$ \\
\midrule
Lexical & RoBERTa & $27.9_{ 6.2}$ & $50.5_{ 8.1}$ & $12.4_{ 2.7}$ & $19.8_{ 13.2}$ & $64.0_{ 22.7}$ & $\underline{12.9_{ 4.2}}$ & $32.9_{ 14.0}$ & $45.3_{ 17.4}$ & $12.6_{ 3.7}$ & $26.8_{ 13.0}$ & $48.4_{ 20.4}$ & $\underline{11.9_{ 4.0}}$ \\
 & $MLM+T_{id}(16)$ & $\underline{30.2_{ 7.0}}$ & $47.3_{ 12.0}$ & $\mathbf{11.9_{ 4.3}}$ & $\mathbf{23.9_{ 13.1}}$ & $52.5_{ 28.6}$ & $13.5_{ 6.0}$ & $31.3_{ 16.1}$ & $50.1_{ 25.6}$ & $\mathbf{11.4_{ 5.7}}$ & $\mathbf{31.4_{ 9.1}}$ & $39.6_{ 13.2}$ & $\mathbf{10.9_{ 3.1}}$ \\
 & $+T_h$ & $29.2_{ 6.4}$ & $47.3_{ 15.3}$ & $\underline{12.2_{ 4.8}}$ & $19.2_{ 12.0}$ & $62.3_{ 25.4}$ & $\mathbf{12.1_{ 5.1}}$ & $34.8_{ 12.2}$ & $43.2_{ 24.3}$ & $\underline{11.5_{ 6.0}}$ & $30.0_{ 7.9}$ & $35.4_{ 10.6}$ & $12.9_{ 4.1}$ \\
 & $+T_o$ & $\mathbf{30.6_{ 5.8}}$ & $\underline{42.3_{ 8.6}}$ & $13.3_{ 4.3}$ & $\underline{23.1_{ 10.9}}$ & $\mathbf{50.5_{ 25.6}}$ & $14.4_{ 4.4}$ & $\mathbf{35.1_{ 11.3}}$ & $\underline{42.0_{ 19.9}}$ & $12.0_{ 5.1}$ & $\underline{30.4_{ 11.8}}$ & $\underline{34.6_{ 16.4}}$ & $13.5_{ 5.7}$ \\
 & $+T_h+T_o$ & $29.4_{ 5.1}$ & $\mathbf{39.5_{ 8.0}}$ & $15.0_{ 4.0}$ & $22.1_{ 8.4}$ & $\underline{51.7_{ 20.0}}$ & $15.3_{ 5.7}$ & $\underline{35.0_{ 10.2}}$ & $\mathbf{36.4_{ 18.4}}$ & $13.8_{ 4.1}$ & $27.9_{ 10.0}$ & $\mathbf{31.8_{ 10.2}}$ & $15.8_{ 3.8}$ \\
\bottomrule
\end{tabular}
\caption{BASIL sentence-level bias under full-corpus partial auxiliary supervision, by annotation type and outlet. F1 is positive-class F1. Erasure is the false-negative rate; attribution is the false-positive rate. Subscripts give standard deviations across ten folds, each with one fine-tuning seed.}
\label{tab:app-basil-full}
\end{sidewaystable*}

\begin{table*}[t]
\centering
\setlength{\tabcolsep}{2.6pt}
\textbf{Task A: seen targets}\par\smallskip
\resizebox{\linewidth}{!}{%
\begin{tabular}{lccccccccc}
\toprule
Model & Stance F1 $\uparrow$ & Reversal $\downarrow$ & A$\rightarrow$F $\downarrow$ & F$\rightarrow$A $\downarrow$ & Atheism $\uparrow$ & Climate $\uparrow$ & Feminism $\uparrow$ & Hillary $\uparrow$ & Abortion $\uparrow$ \\
\midrule
RoBERTa & $\underline{70.7_{ 0.6}}$ & $18.7_{ 0.7}$ & $17.8_{ 2.1}$ & $\mathbf{20.8_{ 3.7}}$ & $\underline{64.3_{ 8.2}}$ & $\mathbf{44.6_{ 0.3}}$ & $56.3_{ 2.2}$ & $\underline{72.8_{ 3.7}}$ & $\mathbf{59.4_{ 5.1}}$ \\
$MLM+T_{id}(16)$ & $\mathbf{71.4_{ 1.5}}$ & $\mathbf{17.2_{ 1.1}}$ & $15.2_{ 2.8}$ & $\underline{21.8_{ 3.8}}$ & $\mathbf{67.2_{ 6.1}}$ & $44.4_{ 0.8}$ & $\mathbf{57.9_{ 2.0}}$ & $\mathbf{73.2_{ 2.3}}$ & $\underline{59.1_{ 2.8}}$ \\
$+T_h$ & $68.1_{ 1.5}$ & $\underline{17.8_{ 1.0}}$ & $\mathbf{12.4_{ 1.6}}$ & $30.7_{ 7.0}$ & $56.2_{ 4.6}$ & $43.8_{ 1.4}$ & $55.8_{ 2.6}$ & $62.5_{ 7.0}$ & $51.5_{ 6.3}$ \\
$+T_o$ & $69.4_{ 1.4}$ & $18.8_{ 1.0}$ & $16.2_{ 3.6}$ & $25.1_{ 6.0}$ & $54.4_{ 12.7}$ & $\underline{44.5_{ 0.7}}$ & $\underline{56.8_{ 1.5}}$ & $67.7_{ 4.5}$ & $57.2_{ 2.8}$ \\
$+T_h+T_o$ & $69.0_{ 1.7}$ & $18.0_{ 1.6}$ & $\underline{12.8_{ 4.7}}$ & $30.4_{ 6.2}$ & $53.6_{ 12.0}$ & $\mathbf{44.6_{ 0.6}}$ & $55.1_{ 2.8}$ & $62.9_{ 8.7}$ & $54.0_{ 4.9}$ \\
\bottomrule
\end{tabular}%
}
\par\medskip
\textbf{Task B: held-out target (Trump)}\par\smallskip
\begin{tabular}{lcccc}
\toprule
Model & Stance F1 $\uparrow$ & Reversal $\downarrow$ & A$\rightarrow$F $\downarrow$ & F$\rightarrow$A $\downarrow$ \\
\midrule
RoBERTa & $19.9_{ 6.9}$ & $\mathbf{9.8_{ 3.7}}$ & $\mathbf{0.5_{ 0.6}}$ & $\mathbf{28.6_{ 11.2}}$ \\
$MLM+T_{id}(16)$ & $25.7_{ 5.2}$ & $\underline{15.3_{ 3.9}}$ & $2.2_{ 1.8}$ & $\underline{41.6_{ 11.0}}$ \\
$+T_h$ & $\underline{27.4_{ 3.0}}$ & $28.4_{ 3.7}$ & $1.9_{ 1.5}$ & $81.9_{ 12.9}$ \\
$+T_o$ & $26.5_{ 8.4}$ & $15.4_{ 6.4}$ & $2.1_{ 1.8}$ & $42.3_{ 20.4}$ \\
$+T_h+T_o$ & $\mathbf{29.9_{ 6.6}}$ & $20.9_{ 4.8}$ & $\underline{1.7_{ 0.9}}$ & $59.9_{ 14.6}$ \\
\bottomrule
\end{tabular}
\caption{SemEval target-conditioned stance under full-corpus partial auxiliary supervision. Stance F1 is the mean of Against and Favor F1. Reversal is endpoint-to-endpoint error, split into its two directions, where A and F denote Against and Favor. Task A also reports stance F1 by target. Subscripts give standard deviations across five fine-tuning seeds.}
\label{tab:app-semeval-full}
\end{table*}

\end{document}